\documentclass{article}

\usepackage[utf8]{inputenc} 
\usepackage{hyperref}

\usepackage{xcolor}
\definecolor{darkblue}{RGB}{0,0,139}

\hypersetup{
  colorlinks=true,            
  citecolor=darkblue,         
  linkcolor=darkblue,         
  urlcolor=darkblue,          
  pdfborderstyle={/S/U/W 1},  
}

\usepackage[preprint]{neurips_2025}

\usepackage{url}
\usepackage{booktabs}       
\usepackage{amsfonts}       
\usepackage{nicefrac}       
\usepackage{microtype}      
\usepackage{lipsum}		
\usepackage{graphicx}
\usepackage{natbib}
\usepackage{doi}
\usepackage{amsmath} 
\usepackage{todonotes}
\usepackage{graphicx}
\usepackage{subcaption}
\usepackage{float}
\usepackage{wrapfig}
\usepackage{breakurl}

\author{%
  Hans Gundlach \thanks{Corresponding author: \texttt{hansgund@mit.edu}} \\
  MIT CSAIL \\
  \texttt{hansgund@mit.edu}
  \And
  Jayson Lynch \\
  MIT CSAIL \\
  \texttt{jaysonl@mit.edu}
    \And
  Neil Thompson \\
  MIT CSAIL \\
  \texttt{neilt@mit.edu}
}

\begin{document}

\title{Meek Models Shall Inherit the Earth}
\maketitle



\begin{abstract}

The past decade has seen incredible scaling of AI systems by a few companies, leading to inequality in AI model performance.  This paper argues that, contrary to prevailing intuition, the diminishing returns to compute scaling will lead to a convergence of AI model capabilities. In other words, meek models (those with limited computation budget) shall inherit the earth, approaching the performance level of the best models overall. We develop a model illustrating that under a fixed-distribution next-token objective, the marginal capability returns to raw compute shrink substantially. Given current scaling practices, we argue that these diminishing returns are strong enough that even companies that can scale their models exponentially faster than other organizations will eventually have little advantage in capabilities.  As part of our argument, we give several reasons that proxies like training loss differences capture important capability measures using evidence from benchmark data and theoretical performance models. In addition, we analyze empirical data on the capability difference of AI models over time. Finally, in light of the increasing ability of meek models, we argue that AI strategy and policy require reexamination, and we outline the areas this shift will affect.

\end{abstract}



\section{Introduction}
Artificial intelligence systems have grown considerably in the last decade \citep{Sevilla_2022}.  This trend has drastically changed the landscape of machine learning. Large corporations now dominate the training of many state-of-the-art (SOTA) models, including systems such as GPT \citep{brown2020language}, Llama \citep{touvron2023llama}, and Gemini \citep{google2023gemini}. Further, it is getting harder to run inference on these models, which now involves the use of multiple GPUs for the largest systems. 

What do these trends mean for the effects of AI on society? If model investment growth continues in this direction, only centralized entities such as the government and corporations can train and use these AI systems \citep{cottier2024rising}. At the same time, other sources warn about the diminishing returns to AI scaling \citep{lohn2023scaling,lu2025race,thompson2021diminishing}. These have raised speculation that AI is ``hitting a wall'' \citep{caputo2025governing}. In this paper, we want to outline models of how performance inequality develops between deep learning models. This model leads us to the counterintuitive conclusion that relevant AI performance levels could converge under the current AI scaling paradigm. Hence, models trained or run with limited resources ``meek'' models, will have more comparable performance to state-of-the-art models. We argue that this could imply greater democratization of AI systems and lead to a world where \textbf{meek models shall inherit the earth}.







\section{Modeling Training Inequality}
\begin{figure}[h]
  \centering
  \includegraphics[width=0.75\textwidth]{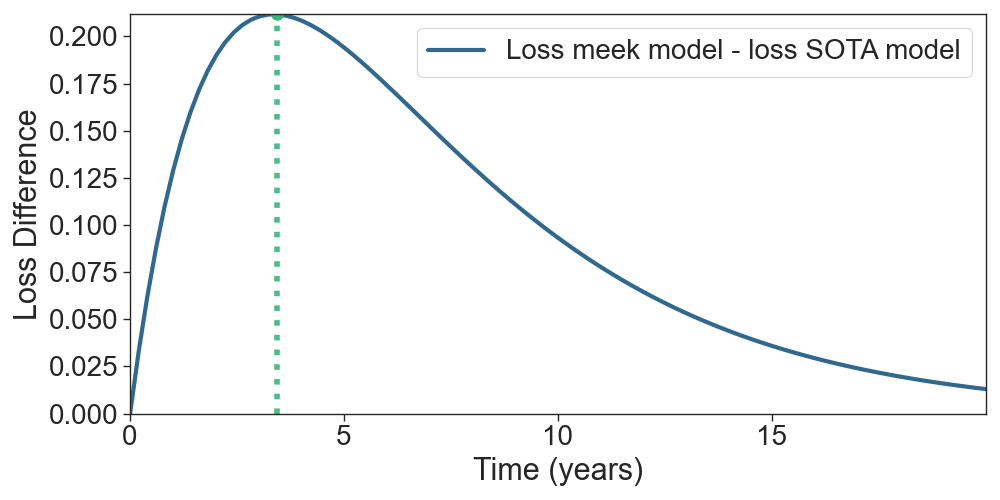}
  \caption{Graph of loss Difference between a model with 3.6x yearly compute scaling and a meek model with a constant compute budget ($\$1000 $ budget). Both models start with an initial compute budget of $\$1000 $. Initially, the model with an exponentially growing compute budget is able to surpass a model created with a constant training budget. However, this gap eventually declines as the top model faces decreasing returns to compute scaling.}
  \label{fig: Training Inflection}
\end{figure}
The first model we construct focuses on the difference in training loss between SOTA models and a ``meek'' model trained with a fixed capital training budget on the same data distribution (we assume $\$1000 $ dollars at $ \approx 10^{17}$ GPU Flops per dollar \citep{li2021a100tco} training budget). We assume that scaling performance is governed by chinchilla-like scaling laws \citep{hoffmann2022trainingcomputeoptimallargelanguage}. Chinchilla laws give a relationship between optimal compute usage $C$ and the log-likelihood loss $L$. We will refer to this as simply the loss for the rest of the paper.


The general form for this relationship, as well as its coefficients are given by the following equations \citep{hoffmann2022trainingcomputeoptimallargelanguage,pearce2024reconciling}.
\begin{align}
L_{opt}(C) &= 1070 \; C^{-0.155} + 1.7 \\
L_{opt}(C) &= A\; C^{-\alpha} + L_{0}
\end{align}  
Over time, the amount of effective Flops per dollar increases. This change is due to two effects. 

The first effect is an increase in hardware capability. This effect is based on trends like Moore's law where the number of transistors on integrated circuits increases by approximately a factor of 2 every 2 years. Our models assume hardware growth $g_h=1.4$ consistent with Moore's law and trends in GPU price performance for GPUs used in ML research \citep{epoch2022trendsingpupriceperformance,rupp2015microprocessor}.




The second effect is due to algorithmic progress, the fact that better algorithms make it possible to learn more effectively with less computation. For example, the discovery of transformers made it possible to train AI models much more effectively in parallel. \citet{ho2024algorithmicprogresslanguagemodels} discovered that algorithmic progress in language models is remarkably rapid consistent over time. Due to algorithmic progress, effective computational resources double approximately every 8 months. We label the growth rate in the effective computation in one year due to algorithmic progress $g_{alg}= 2.8$ \citep{ho2024algorithmicprogresslanguagemodels}. 

We label the compute budget in Flops at time $t=0$ as $C_0$. Therefore, the effective computation resources over time at a given budget $C_0$ is $(g_{alg} g_h)^t C_0$.  To account for the compute budget of large corporations and the progress of SOTA models we must consider a third factor -- the growth rate in compute investment. Compute usage in language models has also grown at a steady exponential rate for the largest models. Between 2010 and 2022 the compute used for training deep learning model grew by a factor of 5 yearly \citep{Sevilla_2022}. We then divide the compute growth rate by the hardware growth rate to get the growth rate in compute investment $g_{i}= 5/1.4 = 3.57$.






Now we have an expression for compute over time, we can find the difference in loss between theoretical SOTA models with exponential growth in investment and ``meek'' individuals with a constant ($\$1000$ compute budget). 
\begin{multline}
  \text{Training Loss Difference} = \text{Loss Meek} - \text{Loss SOTA}\\
 =  A((g_{alg} g_h)^t C_{0})^{-\alpha} - A((g_{alg} g_h g_i)^t C_0)^{-\alpha}  
\end{multline}

Figure~\ref{fig: Training Inflection} shows a graph of this relationship over time.


\paragraph*{The Inflection Point in Training Loss Advantage}
An important point to note about Figure \ref{fig: Training Inflection} is the inflection point in the training advantage curve. At a critical point in time, the diminishing returns to compute scale in addition to the exponential growth in the shared factors of algorithmic and hardware progress, overwhelm the large model provider's exponentially growing compute budget. At this point in time in our model, increasing investment is only able to create a narrow loss advantage over models trained with a very modest budget. Setting the derivative of the loss difference equal to zero lets us solve for the inflection time as:
\begin{equation}
\begin{split}
   \text{Training Inflection Time} = \frac{1}{\alpha \ln g_i} \left[ \ln \left( \frac{\ln(g_{h} g_{alg} g_i)}{\ln \left(g_h g_{alg}\right)} \right) \right]
\end{split}
\end{equation}
It is important to note that this inflection time is the time since a training budget of $C_0$ was state-of-the-art. GPT-2 was trained in 2019 for around $\$25,000-\$50,000$ \citep{umatechnology2025gpt2cost}, with this baseline, our model would predict a peak SOTA advantage in the early-mid 2020s over models with this training budget. If we model the case where $C_0=\$1000$, then $t=0$ corresponds to the year 2017. Using Pre-chinchilla compute scaling where $L-L_0 \propto C_T^{-0.057}$ \citep{kaplan2020scalinglawsneurallanguage}, the inflection time is significantly longer at about 10 years. Trends in AI resource scaling may change significantly. We also believe there are other reasonable parameter choices. Our model conclusions do not depend significantly on these variations (see Fig~\ref{fig:growth_rate_variation} and Fig~\ref{fig:no_mote_capital}).

\subsection{Inference Time Scaling}
We are in the middle of a transition from scaling training compute to scaling inference computation \citep{you2025howfar}. We might not care if we cannot run state of the art models if we get the same result leveraging cheap models with significant inference compute. Will this new paradigm eventually yield the same diminishing returns as pretraining compute? Under the popular model of inference scaling proposed by \citet{epoch2023tradingoffcomputeintrainingandinference}, our results remain valid.\citet{epoch2023tradingoffcomputeintrainingandinference} hypothesizes that inference compute can substitute for and multiply training compute. This means doubling training compute while halving inference compute leads to the same level of performance. In this case, exponentially increasing inference compute would lead to similar diminishing returns in terms of loss and benchmark performance. \citet{you2025howfar} already projects progress in reasoning models to slow significantly. However, significant inference compute might yield new types of pretraining and inference compute might be qualitatively different (see Section~\ref{sec:counterarguments}).


\section{Does Loss Difference Actually Capture Something Important?}
\label{sec:Actually Capture Something Important?}
Model loss has been the focus of neural scaling laws and relates to how well the model predicts the distribution of data it was trained on. However, we are actually interested in the capabilities and usefulness of our models, which may not be the same as the ability to predict portions of a large corpus of text. In this section we discuss loss in more detail and why it might relate to what we would ideally like to measure.

The most straightforward interpretation of loss difference is as a measure of how much better one model is able to predict text than another. Language models loss is traditionally given as the average negative log-likelihood loss per token on a given test set. Another common metric is Perplexity, which is 2 (or $e$ if measured in nats) to the power of the negative log-likelihood loss. More formally, given two models with loss $L_1$ and $L_2$, the difference $L_1-L_2 = \Delta L$ gives the number of extra nats/bits per token necessary for one model needs to encode text over the other.





 Another reasonable perspective is to look at historical trends in loss as a measure. The loss of GPT-3 (davinci) is 4.36  while the loss of GPT-2 (large) is 5.16 \citep{ho2024algorithmicprogresslanguagemodels}.  Since the beginning of deep learning, loss has decreased steadily and tracked AI progress \citep{ho2024algorithmicprogresslanguagemodels}. However, this loss might not correspond to general model capabilities or intelligence. Yet, more tractable metrics like image classification accuracy for vision transformers have a remarkably similar power-law form to this log-likelihood loss (see equation ~\ref{eq:image_equat}) \citep{zhai2022scalingvisiontransformers}. Our analysis holds with this power law formulation as well. 

\begin{align}\label{eq:image_equat}
\min_{N, D} L &= 0.09 + \frac {0.26}{(C + 0.01)^{0.35}}
\end{align}

We speculate that most model capabilities are monotonic functions of loss. Some models' capabilities are accurately captured by such scaling laws while other capabilities are better modeled as abrupt discontinuities. We extend our analysis to these capabilities in Section~\ref{sec:Loss to Benchmark Performance}.
However in less circumscribed or competitive setting, loss difference might not capture the relative performance difference between models, see Section~\ref{sec:counterarguments}.



\subsection{Loss to Benchmark Performance}\label{sec:Loss to Benchmark Performance}
\begin{figure*}[!t]
\centering
\begin{minipage}[t]{0.49\textwidth}
\vspace{0pt}
  \centering
  \includegraphics[width=\linewidth]{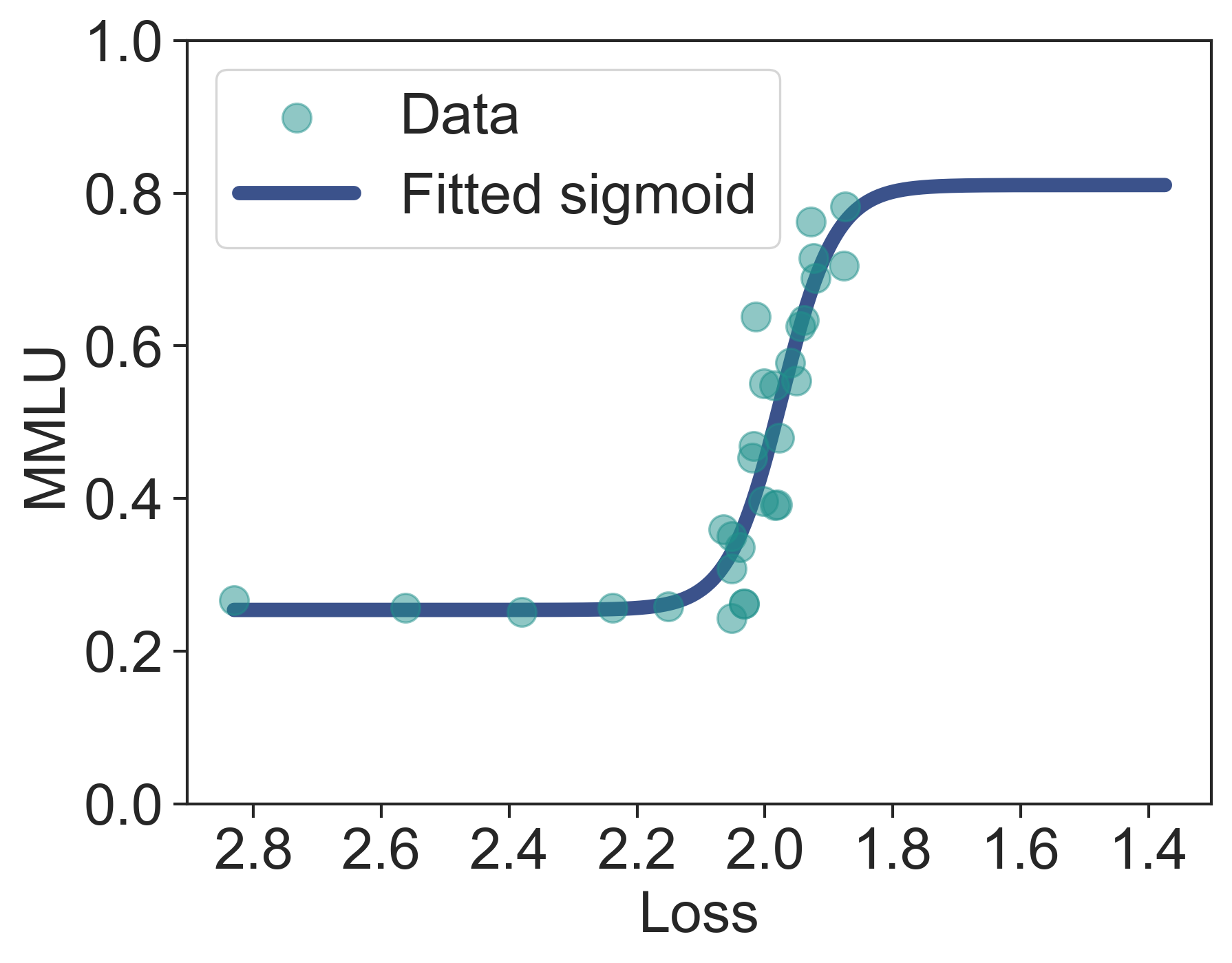}
  \caption{Sigmoid fit of MMLU benchmark performance vs inferred loss.}
  \label{fig:Loss_vs_MMLU}
\end{minipage}%
\hfill
\begin{minipage}[t]{0.49\textwidth}
\vspace{0pt}
  \centering
  \includegraphics[width=\linewidth]{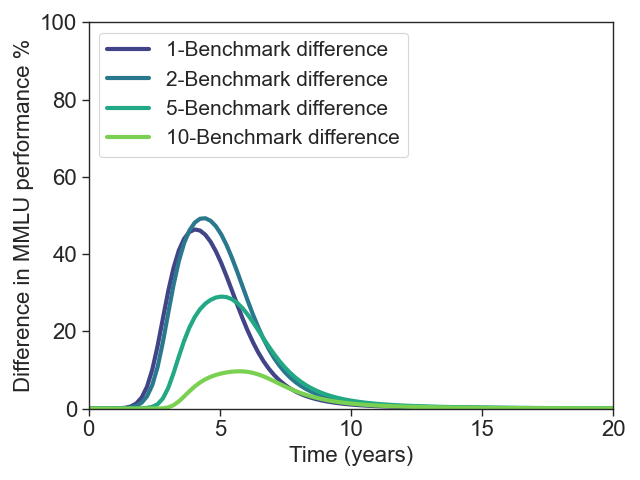}
    \caption{Difference in MMLU performance between SOTA and meek model. 2 and 5 benchmark performance identify difference in capability in tasks which involve multiple (2 to 5) correct MMLU answers.
    The difference in height is a direct result of the $80\%$ sigmoid fit ceiling; a higher ceiling would yield a different height relationship. }
    \label{fig:benchmark_trend_curves}
\end{minipage}
\end{figure*}

How does loss measure actual capabilities? We can take benchmark performance as a proxy for capabilities and look at the relationship between LLM loss and standard benchmarks. Other work has examined this relationship with respect to the 
MMLU (Massive Multitask Language Understanding) and showed the benchmark performance can be modeled as a sigmoid of training compute \citep{owen2024predictablelanguagemodelbenchmark}. The often described slow progress and then rapid improvements in benchmarks before saturating suggest sigmoids should apply to this relation more broadly.  Since loss is a monotonically decreasing function of compute and sigmoids are a strictly monotonic function we can translate this performance and use it as a proxy. 
In Figure~\ref{fig:Loss_vs_MMLU}, we make an updated sigmoid fit of MMLU scores to loss. We determined the loss for each model using Chinchilla scaling laws from data on parameters, and data for each model from \citet{owen2024predictablelanguagemodelbenchmark}.
\begin{align}
\text{Benchmark-Performance} &= \frac{A}{1 + e^{-k(L - x_0)}} + b
\end{align}

Using this sigmoid-translated loss has similar dynamics to those we have previously outlined between SOTA and meek models. At first there is little difference in capabilities, then a large difference emerges, followed by an eventual convergence. We can also consider the case where an overall task requires $p$-steps where each step requires correct zero-shot benchmark performance, which can be modeled as individual benchmark performance to the power of $p$.
\begin{align}
\text{p-Benchmark-Performance} &= \left( \frac{A}{1 + e^{-k(L - x_0)}} + b \right) ^p
\end{align}

In this case, we get an interesting relationship as seen in Figure~\ref{fig:benchmark_trend_curves}. 
As the number of necessary tasks increases, so does the length of time large model builders have an advantage. The maximum loss difference decreases as well. This is due to our fit and the nature of the MMLU benchmark. No model in our dataset have MMLU performance above $80\%$. Therefore, the maximum performance for a p-level task is $0.8^p$. With high accuracy tasks, this effect would be less pronounced. 



\subsection{Hypothesis Test View}\label{Short Hypothesis Test View}


Benchmark performance is a reasonable proxy for some capabilities. However, many AI benchmarks are close to saturation \citep{ott2022mapping}. Benchmarks have a performance threshold that they cannot exceed. Convergence on these tasks might not reflect general intelligence differentials but rather the circumscribed nature of the task. Here we present a more theoretical information theory-based perspective using the assumptions of \citet{epoch2023thedirectapproach}, which parallels our main conclusion. This approach considers two models, A and B, which try to model some base text sampled from a distribution $p_0$, and we ask how many tokens $N$ it takes an ideal observer to distinguish which model is better. In this case, $p_0$ corresponds to the distribution of human text, which we model using common assumptions as stationary and ergodic \citep{jurafsky2025speech3ed}. The expected number of tokens needed to differentiate the two models is given by equation~\ref{eq:token_diff}, where $\alpha$ is the probability that we reject the true hypothesis  (i.e, the test concludes that model A predicts the distribution better than model B, while the reverse is true and vice versa). We set this at $5 \%$

\begin{equation}\label{eq:token_diff}
E_{p_0}\!\left[ N \right]
  \;=\;
  \frac{%
    \left(1-\alpha\right)\,
    \log\!\left(\frac{1-\alpha}{\alpha}\right)
    \;+\;
    \alpha\,
    \log\!\left(\frac{\alpha}{1-\alpha}\right)
  }{\Delta L}
\end{equation}
Equation~\ref{eq:token_diff} shows that as the loss difference decreases, the number of tokens necessary for discrimination increases. An exponential decrease in loss leads to an exponential increase in necessary discriminator tokens.  
Figure~\ref{fig:model_dynamics_threshold} shows the growth in the number of discrimination tokens necessary, increasing over time. Initially, fewer tokens are needed to differentiate models. However, due to the shrinking loss difference, the number of tokens necessary to differentiate the meek model from the SOTA model increases gradually as the two models become more indistinguishable. This dynamic supports the conclusion using our other approaches. \footnote{We must note that this is the expected number of discrimination tokens using a random natural language sample. Fewer discrimination tokens would be needed if we narrow the focus to specialized knowledge. }




\section{Modeling Zero-Shot Inference Inequality}\label{Modeling Zero-Shot Inference Inequality}


In many cases, individuals do not want to train their own models but simply want to run inference on a state of the art model. Therefore, we want to model the difference between the performance of models run with a fixed inference budget vs the performance of a state-of-the-art model. Here, we present a model to do this. Let us consider a meek user with an inference budget $C_{inf}=\$10^{-8}/\text{token}$, which we compare against the same SOTA models as before. Inference costs are primarily driven by the number of parameters. We can find the effective model size we can run by decomposing the inference price into three components Flops/dollar, which is governed by Moore's law. The number of parameters executed per flop, which is influenced by innovations like KV-caching and sparse attention \citep{zhao2024alisa}. And finally, the effective parameters per actual parameters which models innovations like speculative sampling \citep{chen2023accelerating}, distillation, overtraining, and training algorithmic progress which affects effective parameters (see \citet{ho2024algorithmicprogresslanguagemodels})


\begin{equation}
P_{\text{eff}}(t) = C_{\text{inf}} (\$) \underbrace{\left[\frac{\text{FLOP}}{\$}(t)\right]}_{g_{h}} \underbrace{\left[\frac{P}{\text{Flop}}(t)\right]\left[\frac{P_{\text{eff}}}{P}(t)\right]}_{g_{inf}}.
\end{equation}
The extent of algorithmic progress in LLM inference ($g_{inf}$) is still an open question. However, to get a rough numerical estimate, we use trends in the cost of inference by \citet{epoch2025llminferencepricetrends}. We use the slowest estimate of 9x per year to get a more conservative picture, as well as to limit the interference of competitive rather than technical effects. This trend can be decomposed as 
\begin{equation}
    \frac{\$}{\text{token}} = \left[\frac{P_{\text{eff}}}{\text{token}} \right] \left[\frac{P}{P_{\text{eff}}} \right] \left[\frac{FLOP}{P} \right] \left[\frac{\$}{FLOP} \right]
\end{equation}

We do not know of any factors which effect  $P_{\text{eff}}/\text{token}$ so we assume it is constant. 
Therefore, we can directly identify the inverse of these  price changes with the factors that affect $P_{\text{eff}}(t)$. The effective model size we can run thus increases 9x per year. What is the performance of a model of size $P_{\text{eff}}$? We assume that this effective model corresponds to a model that is trained chinchilla optimally. For example, we could perform distillation where the large teacher model is trained chinchilla optimally, but the actual number of parameters used in practice is closer to the student model.


Therefore, the effective compute of this model at time t is $\approx g_{alg}^{t}P_{\text{eff}}(t)^2$. Technically, $g_{alg}= g_{N}+g_{D}$ where $g_{N}$ is the effect of better training on effective parameters, while $g_{D}$ is the increase in effective data size due to algorithmic progress (see \citet{ho2024algorithmicprogresslanguagemodels}). This means we are overestimating because $P_{\text{eff}}$ also takes into account this growth. However, the baseline empirical measures $g_{N}$ are an order of magnitude less than $g_{D}$ \citep{ho2024algorithmicprogresslanguagemodels} or the growth of inference efficiency we see in \citet{epoch2025llminferencepricetrends} so we neglect this term. Hence, we can model the meek inference loss as: 

\begin{align}
\text{Meek Inference Loss} &= \max(L_0 + A((g_h^t g_{inf}^tC_{inf})^2 g_{alg}^{t})^{-\alpha}, \text{SOTA Loss})
\end{align}
Therefore, the loss difference between a meek inference model and an SOTA model is : 
\begin{multline}
\max(\text{SOTA Loss-Meek Inference Loss}, 0)\\ =  \max(A((g_{alg} g_h g_i)^t C_{0})^{-\alpha} - A((g_h^t g_{inf}^tC_{inf})^2 g_{alg}^{t})^{-\alpha}, 0) 
\end{multline}
This formula demonstrates the extremely strong effect of inference improvements. If a user is able to cut inference cost by a factor of 10 they can run a model that is 10 times larger. This 10x larger model corresponds to a model that is trained with 100x the amount of compute as used in the smaller model. However, an individual cannot do this indefinitely. Eventually, they are able to run the largest SOTA model and therefore cannot gain any more loss-performance benefit.

\begin{figure}[h!]
  \centering
  \includegraphics[width=0.7\textwidth]{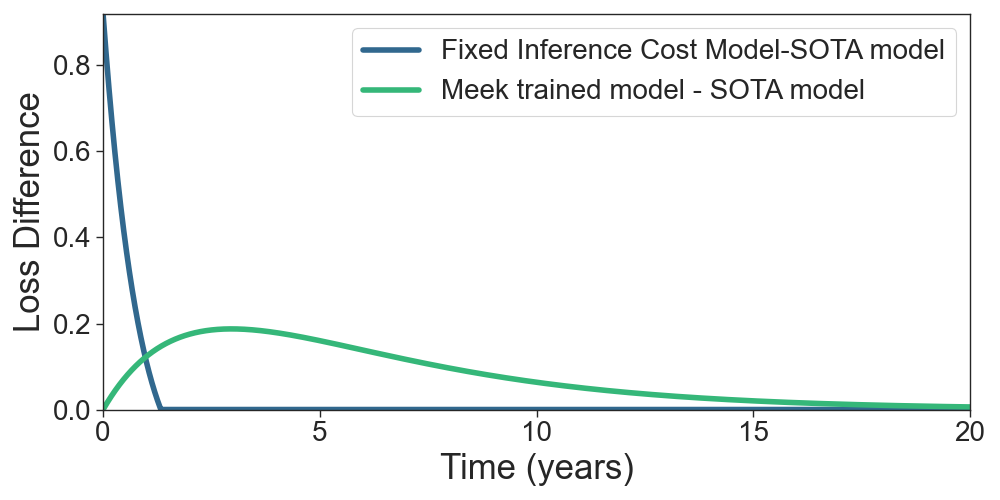}
  \caption{Graph of loss difference in inference vs training performance. The inference difference is between a SOTA model and a model that can be run with an example fixed inference budget. For comparison, we have the training loss difference between the SOTA model and the meek model with a fixed training budget. Notice the much quicker drop in loss difference under our inference model.}
  \label{fig:inf_differential}
\end{figure}

\section{Empirical Trends}
\begin{figure*}[h!]
    \centering
    \includegraphics[width=.9\linewidth]{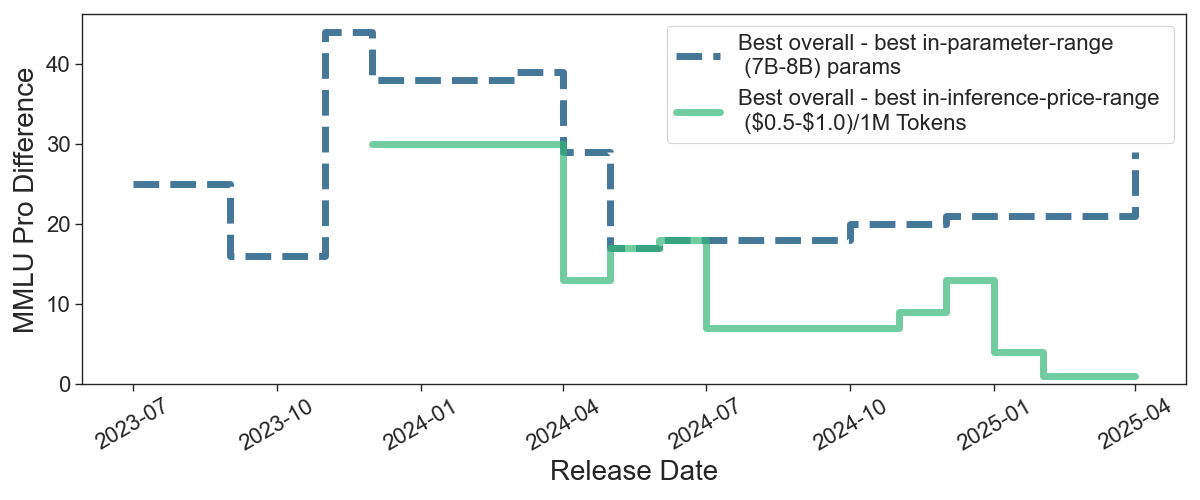}
    \caption{Graph depicting the difference in MMLU-Pro score between the model with the maximum score overall and the best score among models within a fixed inference price range $0.5\$-1\$$ per 1M tokens. Models sourced from the Artificial Analysis LLM Leaderboard \cite{artificialanalysis2025}.}
    \label{fig:mmlu_param_count_comparison.png}
\end{figure*}

How has AI inequality actually developed in recent years? The data available is quite sparse but still suggestive of interesting developments in line with our model. In our case, we use data sourced from the Artificial Analysis LLM Leaderboard \citep{artificialanalysis2025}, which tracks metrics on commercially available LLMs. As in our inference model, we compare the best performance of models with a fixed inference budget here $\$0.5-\$1/1M \text{ tokens}$ to the best performing models overall. Current models leverage significant scaffolding and reasoning levels, which makes comparison difficult. We would expect that the best systems available for commercial use would leverage higher inference costs for more extensive reasoning and hence perform much better. Despite this, we see a remarkable convergence of model capabilities in line with the fast convergence we see in our zero-shot inference advantage model (see Fig~\ref{fig:inf_differential}). We suspect that this is due to fast algorithmic progress in inference as well as the compounded growth available for inference computation (see Section~\ref{Modeling Zero-Shot Inference Inequality}). 
\\
\\
Since we have no known estimate of training compute, we rely on parameter count as a proxy for total training compute. We recognize that this is a very rough measure. Parameter count can be used as a proxy for models that are trained Chinchilla optimally \citep{hoffmann2022training}. However, recent models could be overtrained or use distillation to reduce parameter count and improve inference performance. Further, parameter counts are only available for open-source models. Overall, the data looks inconclusive. Tentatively, based on parameter count, it appears the performance difference between models is constant or diverging. For 7B models (rather than models $1000\$$) the relevant start date would be around 2023 (the date when 7B models were SOTA) \citep{EpochNotableModels2024} with a prediction inflection point around 2027. Therefore, the current MMLU Pro gap would be in line with our predictions. Yet the largest observed gap in MMLU Pro scores—visible in Fig. \ref{fig:mmlu_param_count_comparison.png}, which occurs during 2023 – 2024—falls outside the trajectory forecast by our training-based predictions, provided it represents a genuine trend.



\section{Alternative Views: New Training Paradigms and Adversarial Settings}\label{sec:counterarguments}
We explain this convergence as models growing increasingly close to the fixed distribution of human text. There are only so many abilities necessary to predict human text, and as models grow larger, they master narrower, less common abilities (see \citep{michaud2024quantizationmodelneuralscaling}). This is likely true under naive inference scaling as well. However, powerful AIs can do more than just human imitation. They can far exceed humans and learn new kinds of skills. RL and synthetic data techniques promise to change drastically the distributions learned by AIs. It is no longer a question of how well AIs are learning but what they are learning. We are uncertain about our model in these new cases where AIs are trained on adaptively chosen data or their own synthetic data. Further, there are instances where exponentially small differences in overall loss may correspond to large capability differentials. For instance, in order to model the small set of tokens dealing with elementary math the model has to internalize the rules of arithmetic \citep{michaud2024quantizationmodelneuralscaling}. This is particularly the case in adversarial settings where agents are incentivized to win by finding situations where the competitor is unfamiliar. In competitive games, adversaries with exponentially increasing compute might continuously diverge from their competitor \citep{jones2021scalingscalinglawsboard}. This could be the case because small performance differences make a large impact on win rate success (think of track running for instance).  Yet, diminishing returns may return at high levels of compute as agents approach perfect play \citep{neumann2022scaling}.


\section{AI Governance Discussion}\label{sec:AI Policy Implications}

\begin{figure*}[h!]
\centering
\begin{minipage}[t]{0.49\textwidth}
\vspace{0pt}
  \centering
   \includegraphics[width=\linewidth]{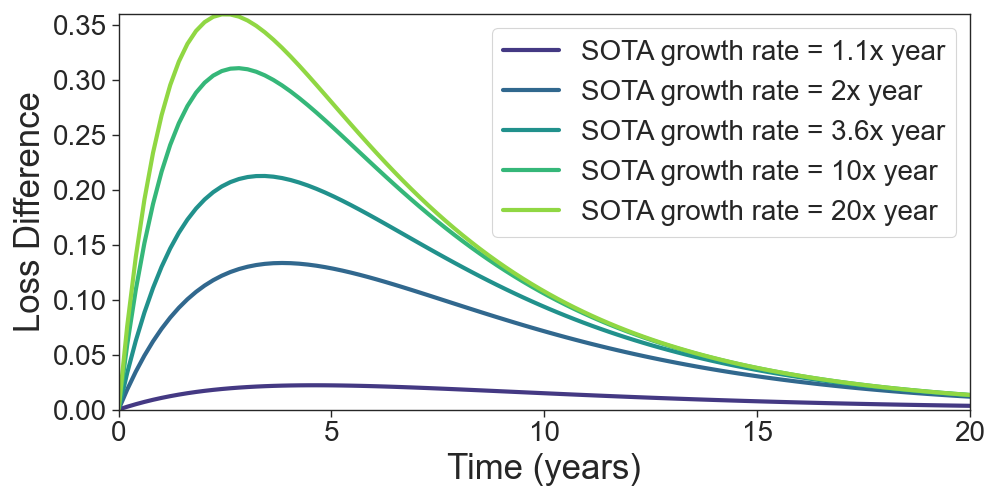}
    \caption{Loss difference between SOTA vs meek models with different levels of SOTA compute investment growth $g_i$ (3.6 is the default).}
    \label{fig:growth_rate_variation}
\end{minipage}%
\hfill
\begin{minipage}[t]{0.49\textwidth}
\vspace{0pt}
\vspace{0pt}
  \centering
  \includegraphics[width=\linewidth]{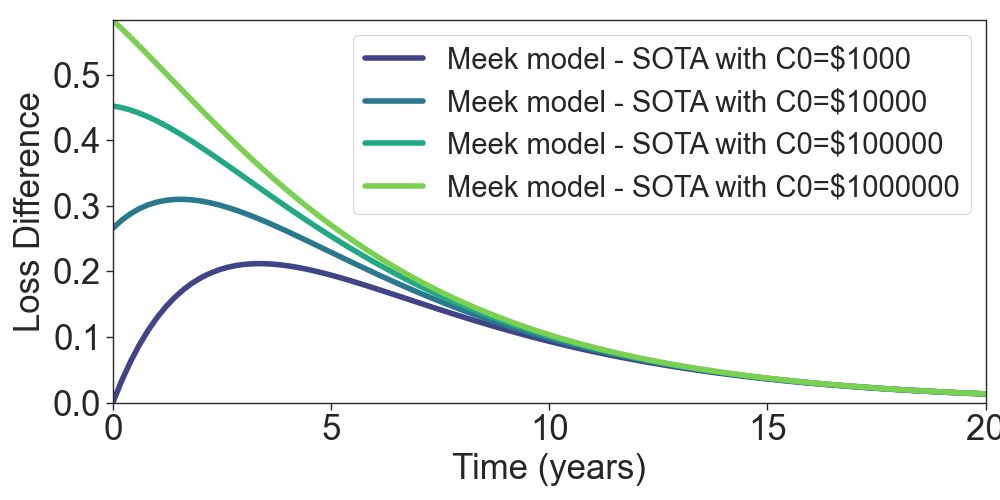}
    \caption{Initial compute capital advantage makes little difference in loss over time. Meek model training budget is kept constant at $\$1000$.}
    \label{fig:no_mote_capital}
\end{minipage}
\end{figure*}

Our model points to several conclusions for AI governance. 
First, there might exist a crucial \textbf{``governance window''} where large entities have a large advantage over ubiquitous AI models. During this period, regulations can be more targeted. This period is particularly advantageous as trusted organizations can gain experience with powerful models and design safety procedures for these models before they become ubiquitous. 
However, it is a double-edged sword, as it allows small groups to accumulate power and influence during this period. 

\paragraph{Can increased AI investment help AI safety?} If the benefits of centralization and using the ``governance window'' are strong enough, accelerating AI might have a positive effect on safety. Appendix~\ref{Robustness and Variation Section} Fig~\ref{fig:growth_rate_variation} depicts the loss advantage with different SOTA investment rates. If a safe organization makes a very large investment in AI it will have access to highly capable systems much earlier than consumers. In this case the large AI loss advantage gives an organization much more time to do safety research before these system becomes ubiquitous. 


\paragraph{Is money a long-term moat?} Companies and countries may care about having a competitive advantage by maintaining private/proprietary foundation models. We've seen that even drastic differences in compute investment in these models create only a modest difference in important capability measures. It is possible that in the long run, no entity can hope to keep a large advantage under the current paradigm simply by having more capital.


\paragraph{AI for all} If these trends continue then it seems access to high performance deep learning models will become as ubiquitous as computer ownership. Given that computers are also becoming less expensive and more widely used, it seems reasonably likely that a large fraction of the world's population could have access to powerful deep learning models. This suggests many people will be able to share in the benefits and productivity increases that might come with improved AI and suggests a likely dispersion of power to individuals. However, this poses a risk to safety if individuals can access dangerous capabilities like bioweapons design. 


\subsection{We Need to Rethink AI Regulation}
Much of current AI governance is focused on monitoring and limiting access to large frontier systems. These include US export controls on GPU hardware. The US and EU focus on models trained with above $10^{26}$ and $10^{25}$ Flops, respectively \citep{caputo2025governing}. Our work shows that simply restricting total compute may not suffice to keep frontier AI capability from becoming ubiquitous. 
Future AI governance would either need to drastically increase capacity for monitoring and safeguarding systems or find new targets to be able to effectively limit access to powerful models. These could include regulating data, algorithms, and new research breakthroughs \citep{caputo2025governing}.

\section{Conclusion}



AI training is stretching the limits of data, computation, and energy and continued scaling may slowdown in the near future \citep{epoch2024canaiscalingcontinuethrough2030}. However, even if continued scaling is possible, there are deeper limits to AI progress: in particular, diminishing returns to scale. We've developed a strategic model in which these diminishing returns imply that the relative performance advantage of very large models over ``meek'' models shrinks over time. Our model, therefore, implies that we may end up in a world where frontier AI capabilities are easily accessible even to actors with disadvantages in capital and compute. In other words, a decentralized world where frontier AI is ubiquitously accessible. However, we emphasize that AI is a rapidly changing field. New technical methods are already in development, which could change the way we develop AI. Regardless, we need to prepare ourselves by developing AI governance methods that are effective in such a world -- a world where meek models inherit the earth.

\section{Acknowledgments}
We would like to thank Zachary Brown for his feedback and suggestions.

\bibliographystyle{unsrtnat}   
\bibliography{references}






\appendix

\section{Full Hypothesis Testing Framework}
\label{sec:Full Hypothesis Testing Framework}


Here we outline the hypothesis testing framework introduced in Section~\ref{Short Hypothesis Test View}, which is based on a model developed by \citet{epoch2023thedirectapproach}. This approach asks how many tokens it takes to declare one model better than another using the Sequential Probability Ratio Test (SPRT) \citep{nowak2011sequential}.


\subsection{Sequential Probability Ratio Test (SPRT)}

Our discrimination model is based on the SPRT test \citep{nowak2011sequential} and the framework developed in \citet{epoch2023thedirectapproach}.
We use the same assumptions they make here to show how our results can incorporate their model; however, assessing when all of these conditions fully hold is out of scope for our analysis and deserves future research. Consider two models, A and B, with predictive probabilities $p_{A}$ and $p_{B}$. We continue measuring the model's probability on given tokens until a likelihood threshold is reached, in which case we declare that one model predicts the text better than the other. Here, we derive the test threshold assuming text samples are generated iid from a true distribution $p_0$ and then extend our proof to the stationary and ergodic case, which is a common modeling assumption for language \citep{jurafsky2025speech3ed}. This is a good approximation; however, natural language can depend on arbitrarily far words, which breaks these assumptions. 


\[
X_1, X_2, \dots, X_n {\sim} p_0
\]
Here, we use an information-theoretic interpretation of the loss as the cross entropy between the model's probability distribution and the true distribution $p_0$. 
\begin{equation}
H(p_0,p_A) = H(p_0)+D_{KL}(p_0 \parallel p_A)
\end{equation}
\citet{epoch2023thedirectapproach} identifies $D_{KL}(p_0 \parallel p_A) = C^{-\alpha}$ with the non-irreducible part of the loss function.

$H_{A}$ is the hypothesis that A is the better model. Likewise, $H_{B}$ represents the hypothesis that B is the better model. Let $Z_{i}$ represent the log-likelihood ratio per token given by: $Z_i = \log \frac{p_A(x_i)}{p_B(x_i)}$. The cumulative log likelihood ratio is given by:
\begin{equation}
Z_{N} = \sum_{i=1}^{N} Z_{i} = \log \left(\prod_{i=1}^{N} \frac{p_A(X_i)}{p_B(X_i)}\right), i = 1, 2, \dots
\end{equation}
Next, we choose two threshold values, $A_{th}$ and $B_{th}$. We choose to accept $H_{A}$ if $Z_{N} \geq log(A_{th})$ and we accept $H_{B}$ if $Z_{N} \leq log(B_{th})$. These thresholds can be chosen such that the probability of falsely rejecting $H_{A}$ is $\alpha$ while the probability of falsely rejecting $H_{B}$ is $\beta$. The threshold values that have such properties are \citep{nowak2011sequential}:
\begin{equation}
 \qquad A_{th} = \frac{1-\beta}{\alpha}, B_{th} = \frac{\beta}{1-\alpha}
\end{equation} 
Now, we can use Wald's stopping theorem to find $N$, the expected number of tokens necessary to differentiate the better model. Wald's stopping theorem holds under stationary and ergodic distributions \citep{FrankenLisek1982}. See also \citep{epoch2023thedirectapproach}.
\begin{equation}
E_{p_0}[Z_N] = E_{p_0}[N] E_{p_0}[Z_{i}]
\end{equation}
where
\begin{equation}
E_{p_0}[Z_{i}] = E_{p_0}[\log \frac{p_A(x_i)}{p_B(x_i)}] = 
D_{KL} (p_0 \parallel p_{B}) - D_{KL}( p_0 \parallel p_{A})= \Delta L
\end{equation}
The proof above also works for stationary and ergodic processes as well, due to ergodicity: 
\begin{equation}
\frac{1}{n} Z_n \xrightarrow[n\to\infty]{\text{a.s.}} E_{p_0}[Z_i]
\end{equation}
We can evaluate the expected value of $Z_{N}$.


\begin{equation}
E_{p_0}\!\left[ Z_N \mid H_A \right] 
  = (1-\beta)\,\log A_{th} + \beta\,\log B_{th}
\end{equation}
Under $H_B$ it is:
\begin{equation}
E_{p_0}\!\left[ Z_N \mid H_B \right]  = (1-\alpha)\,\log B_{th} + \alpha\,\log A_{th}
\end{equation}
This means the expected number of tokens necessary to distinguish the two models under each hypothesis is given by: 
\begin{equation}
E_{p_0}\!\left[ N \mid H_A \right] 
  = \frac{(1-\beta)\,\log A_{th} + \beta\,\log B_{th}}{\Delta L}
\end{equation}
\begin{equation}
E_{p_0}\!\left[ N \mid H_B \right] 
  = \frac{(1-\alpha)\,\log B_{th} + \alpha\,\log A_{th}}{|\Delta L|}, \qquad .
\end{equation}
Finally, we set $\alpha=\beta$ because we want symmetrical considerations. Let us consider the case where $H_A$ corresponds to the larger model (ie, it predicts the text better) and $\Delta L>0$ as the default.  We can then write the expected number of discrimination tokens as:
\begin{equation}
E_{p_0}\!\left[ N \right]
  \;=\;
  \frac{%
    \left(1-\alpha\right)\,
    \log\!\left(\frac{1-\alpha}{\alpha}\right)
    \;+\;
    \alpha\,
    \log\!\left(\frac{\alpha}{1-\alpha}\right)
  }{\Delta L}
\end{equation}

\subsection{Bayes Slowdown Factor}
Similar to \citep{epoch2023thedirectapproach} we experiment with a Bayesian slowdown factor to more closely measure practical distinguishability rather than ideal distinguishability. 


\begin{figure*}[!t]
\centering
\begin{minipage}[t]{0.48\textwidth}
\vspace{0pt}
  \centering
  \includegraphics[width=\linewidth]{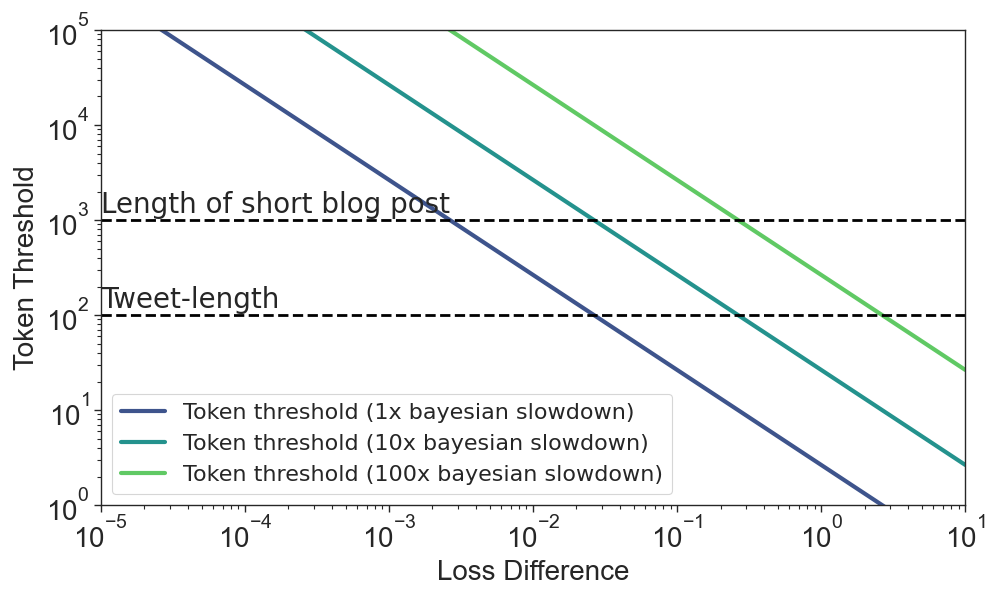}
    \caption{Expected number of tokens needed to differentiate text based on loss using SPRT. Bayesian Slowdown $=1$ corresponds to an ideal discriminator. }
    \label{fig:token_threshold_vs_loss_difference}
\end{minipage}%
\hfill
\begin{minipage}[t]{0.49\textwidth}
\vspace{0pt}
  \centering
  \includegraphics[width=\linewidth]{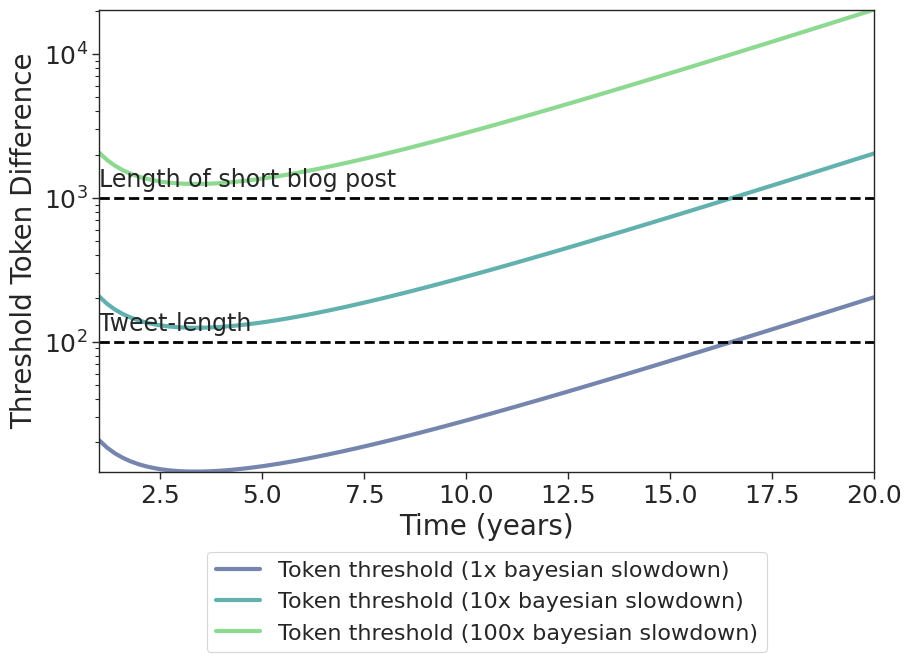}
    \caption{Expected number of tokens needed to differentiate SOTA-model vs meek model using SPRT. Bayesian Slowdown$=1$ represents the number of tokens needed by a perfect discriminator.}
    \label{fig:model_dynamics_threshold}
\end{minipage}
\end{figure*}

\begin{equation}
\log(\text{Posterior odds}) = \log(\text{Prior odds}) + \frac{\log(\text{Bayes factor})}{\text{Slowdown}}.
\end{equation}
Accounting for such a factor scales the number or expected tokens proportionally by the size of the slowdown (see Fig~\ref{fig:token_threshold_vs_loss_difference} and Fig~\ref{fig:model_dynamics_threshold}).

\subsection{Speculative Sampling Implications}
We also think such a model is of independent interest in modeling the gains from speculative sampling. \citep{chen2023accelerating}. Speculative sampling uses a small, cheap draft model to generate most tokens, while a more expensive target model is used for tokens the draft model generates incorrectly. The rise in similarity between models of different sizes motivates the use of techniques like speculative sampling. Our model explains why the target model and draft model are indistinguishable on most tokens. We therefore predict increase usage of techniques like speculative sampling in the future.

\section{Variation in Investment Trend}\label{Robustness and Variation Section}

\begin{figure}[h!]
  \centering
  \begin{subfigure}[t]{0.49\textwidth} 
      \centering
      \includegraphics[width=\textwidth]{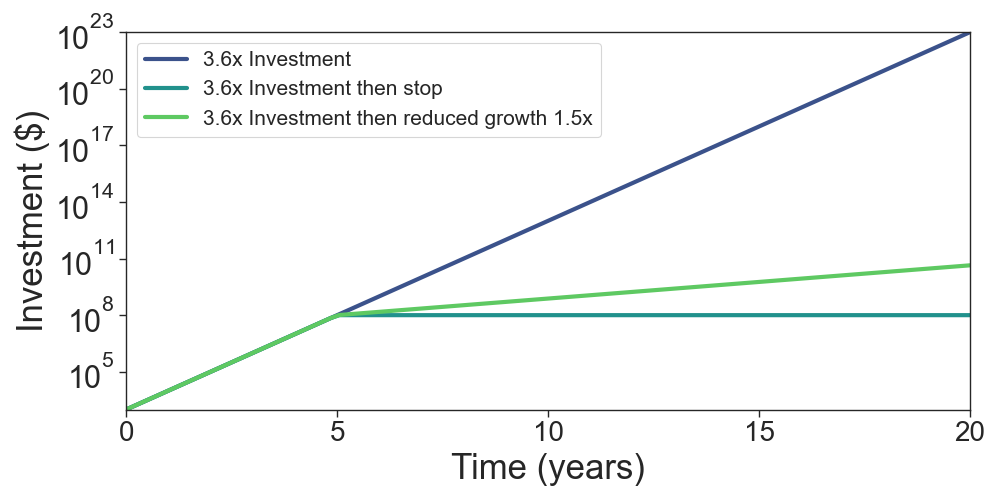}
      \caption{Investment Trends on a semilog graph. }
      \label{fig:investment_trend}
  \end{subfigure}
  \hfill
  \begin{subfigure}[t]{0.49\textwidth} 
      \centering
      \includegraphics[width=\textwidth]{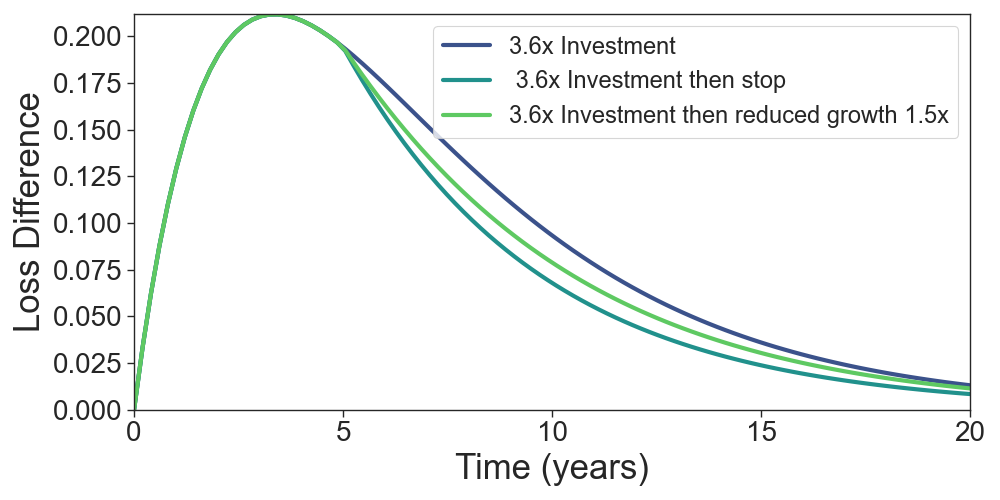}
      \caption{Loss trends based on investment schedules in Fig \ref{fig:investment_trend}}
      \label{fig:second}
  \end{subfigure}
  \caption{Model investment growth trajectories vs loss-difference. Surprisingly, large exponential variation in investment trajectory leads to little change in loss.}
  \label{fig:stagnation_variation}
\end{figure}

The graphs we have presented are based on several key assumptions. These assumptions are that growth in compute investment, algorithms, and hardware will consistently continue. We might be interested in how our model changes in response to AI training growth slowing or coming to a halt after five years. This could be due to energy limits or lack of remaining training data \citep{epoch2024canaiscalingcontinuethrough2030}. Fig~\ref{fig:stagnation_variation} illustrates that such stagnation has little effect on the loss difference in our model, as AI builders are already in a regime with steep diminishing returns to loss.

\end{document}